\documentclass[11pt]{article}
\usepackage{coling2020}
\usepackage{graphicx}
\usepackage{subcaption}
\usepackage{times}
\usepackage{url}
\usepackage{latexsym}
\usepackage{amsmath}
\usepackage[colorinlistoftodos]{todonotes}
\usepackage{graphicx}
\usepackage{booktabs}
\usepackage{multirow}
\usepackage{mathrsfs}
\usepackage{amsmath}
\usepackage{bbm}
\usepackage{amssymb}
\usepackage{wrapfig}
\usepackage{balance}
\usepackage{url}
\usepackage{bm}
\usepackage{tabularx}
\usepackage{float}
\usepackage{amsmath}
\usepackage{amsthm}
\usepackage{amsfonts}
\usepackage{amssymb}
\usepackage{graphicx}
\usepackage{booktabs}
\usepackage{placeins}
\usepackage{multirow}
\usepackage[ruled, linesnumbered]{algorithm2e}
\usepackage{epstopdf}
\usepackage{color}
\usepackage{subcaption}
\usepackage[T1]{fontenc}
\usepackage{amsmath,amsfonts,bm}
\usepackage{balance}
\usepackage{enumitem}
\usepackage{xcolor}
\usepackage{balance}
\usepackage{wrapfig}
\usepackage{relsize}
\usepackage{subcaption}
\usepackage{enumitem}
\usepackage{wrapfig}
\usepackage{graphics}
\usepackage{url}
\usepackage{makecell}
\usepackage{array}
\newcommand{\PreserveBackslash}[1]{\let\temp=\\#1\let\\=\temp}
\newcolumntype{C}[1]{>{\PreserveBackslash\centering}p{#1}}
\newcolumntype{R}[1]{>{\PreserveBackslash\raggedleft}p{#1}}
\newcolumntype{L}[1]{>{\PreserveBackslash\raggedright}p{#1}}
\colingfinalcopy 

\usepackage{hyperref}
\definecolor{darkblue}{rgb}{0, 0, 0.5}
\hypersetup{colorlinks=true,citecolor=darkblue, linkcolor=darkblue, urlcolor=darkblue}

\title{LTIatCMU at SemEval-2020 Task 11: Incorporating Multi-Level Features for Multi-Granular Propaganda Span Identification}

\author{Sopan Khosla\thanks{* denotes equal contribution.} \quad Rishabh Joshi\footnotemark[1] \quad Ritam Dutt\footnotemark[1]  \quad Alan W Black \quad Yulia Tsvetkov\\
Language Technology Institute \\ Carnegie Mellon University \\
\{{\tt sopank,rjoshi2,rdutt,awb,ytsvetko\}@cs.cmu.edu}
}

\date{}

\begin{document}
\newcommand{\refalg}[1]{Algorithm \ref{#1}}
\newcommand{\refeqn}[1]{Equation \ref{#1}}
\newcommand{\reffig}[1]{Figure \ref{#1}}
\newcommand{\reftbl}[1]{Table \ref{#1}}
\newcommand{\refsec}[1]{Section \ref{#1}}

\newcommand{\add}[1]{\textcolor{red}{#1}\typeout{#1}}
\newcommand{\remove}[1]{\sout{#1}\typeout{#1}}

\newcommand{\m}[1]{\mathcal{#1}}
\newcommand{\bmm}[1]{\bm{\mathcal{#1}}}
\newcommand{\real}[1]{\mathbb{R}^{#1}}
\newcommand{\method}{\textsc{MedPrint}}

\newcommand{\bleuone}{BLEU${_1}$\xspace}
\newcommand{\bleu}{BLEU${_4}$\xspace}

\newcommand{\problem}{}
\newcommand{\problemfull}{}

\newtheorem{theorem}{Theorem}[section]
\newtheorem{claim}[theorem]{Claim}

\newcommand{\reminder}[1]{\textcolor{red}{[[ #1 ]]}\typeout{#1}}
\newcommand{\reminderR}[1]{\textcolor{gray}{[[ #1 ]]}\typeout{#1}}

\newcommand{\tensor}{\mathcal{X}}
\newcommand{\Real}{\mathbb{R}}

\newcommand{\tuples}{\mathbb{T}}

\newcommand\norm[1]{\left\lVert#1\right\rVert}

\newcommand{\note}[1]{\textcolor{blue}{#1}}

\newcommand*{\Scale}[2][4]{\scalebox{#1}{$#2$}}%
\newcommand*{\Resize}[2]{\resizebox{#1}{!}{$#2$}}%

\def\mat#1{\mbox{\bf #1}}
\newcommand{\cev}[1]{\reflectbox{\ensuremath{\vec{\reflectbox{\ensuremath{#1}}}}}}

\maketitle
\begin{abstract}
In this paper we describe our submission for the task of Propaganda Span Identification  in news articles. We introduce a BERT-BiLSTM based span-level propaganda classification model that identifies which token spans within the sentence are indicative of propaganda. The ``multi-granular'' model incorporates linguistic knowledge at various levels of text granularity, including word, sentence and document level syntactic, semantic and pragmatic affect features, which significantly improve model performance, compared to its language-agnostic variant. To facilitate better representation learning, we also collect a corpus of 10k news articles, and use it for fine-tuning the model.
The final model is a majority-voting ensemble which learns different  propaganda class boundaries by leveraging different subsets of incorporated knowledge.\footnote{Our final ensemble attains $4^{th}$ position on the test leaderboard. Our final model and code is released at \url{https://github.com/sopu/PropagandaSemEval2020}.}

 \blfootnote{
    %
    \hspace{-0.65cm}  
    This work is licensed under a Creative Commons 
    Attribution 4.0 International License.
    License details:
    \url{http://creativecommons.org/licenses/by/4.0/}
}
\end{abstract}

\section{Introduction}
\label{sec:introduction}
 
Propaganda \cite{bernays1928propaganda} is 
``the deliberate and systematic attempt to shape perceptions, manipulate cognition, and direct behavior to achieve a response that furthers the desired intent of the propagandist'' \cite{jowett2018propaganda}.  
The rise of digital and social media has enabled propaganda campaigns to reach vast audiences~\cite{glowacki2018news,tardaguila2018fake}, influencing or undermining the democratic processes, amplifying the effects of misinformation, and creating echo-chambers. 

As a tactic of manipulation, propaganda is considered most effective when it is veiled: the readers are not able to identify it, but their opinions are shaped according to the propagandist's hidden agenda~\cite{doc-prop-2019}; it is thus very difficult to identify propaganda automatically. 
However, automatic identification and analysis of propaganda in news articles and on social media are essential to understand propaganda at scale and develop approaches to countering it \cite{king2017chinese,starbird_2018,field2018framing}. 

Prior research on propaganda detection focused primarily on identifying propaganda at a document level, due to the dearth of finer-grained labelled data \cite{doc-prop-2017,doc-prop-2019}. This has resulted in classifying \emph{all} news articles from a particular source as propaganda (or hoax, disinformation, etc.), which is often not the case, and which can obfuscate a finer-grained analysis of propagandistic strategies  \cite{propaganda-emnlp-2019,horne2018sampling}. 

Recently, \newcite{propaganda-emnlp-2019} carried out a seminal task of fine-grained propaganda detection and curated a dataset consisting of about 550 news articles. The dataset contains word-span level annotations provided by high-quality professionals along with additional information about the propaganda technique employed in the span. The Span Identification (SI) sub-task of SemEval 2020, Task 11, \cite{Semeval20task11} employs this aforementioned corpus and requires participants to detect propaganda spans in news articles. 

In this paper, we describe our solution to the SI task. We propose a BERT-BiLSTM based multi-granularity model, inspired from \cite{propaganda-emnlp-2019}. The model's parameters are jointly optimized on the tasks of sentence-level and word-level propaganda detection. We fine-tune BERT-BiLSTM end-to-end on the joint-loss to obtain competitive scores on the SI task. Furthermore, we explore the benefits of incorporating word, sentence and document-level syntactic and affective features extracted from dictionaries like LIWC \cite{liwc} and Empath \cite{fast2016empath}. We find that incorporating these features leads to significant improvements in performance (discussed in Section \ref{sec:setup}). We also leverage articles from multiple well-known propaganda news sources to fine-tune BERT 
in an unsupervised manner. Furthermore, in order to tackle the problem of unbalanced data, we use a weighted cross-entropy loss.
We present our model and additional features in Section \ref{sec:details}, and show our results and ablation analysis in Section \ref{sec:setup}.
\begin{figure*}[t]
\centering
\captionsetup{font=small,labelfont=small}
\includegraphics[width=0.8\textwidth]{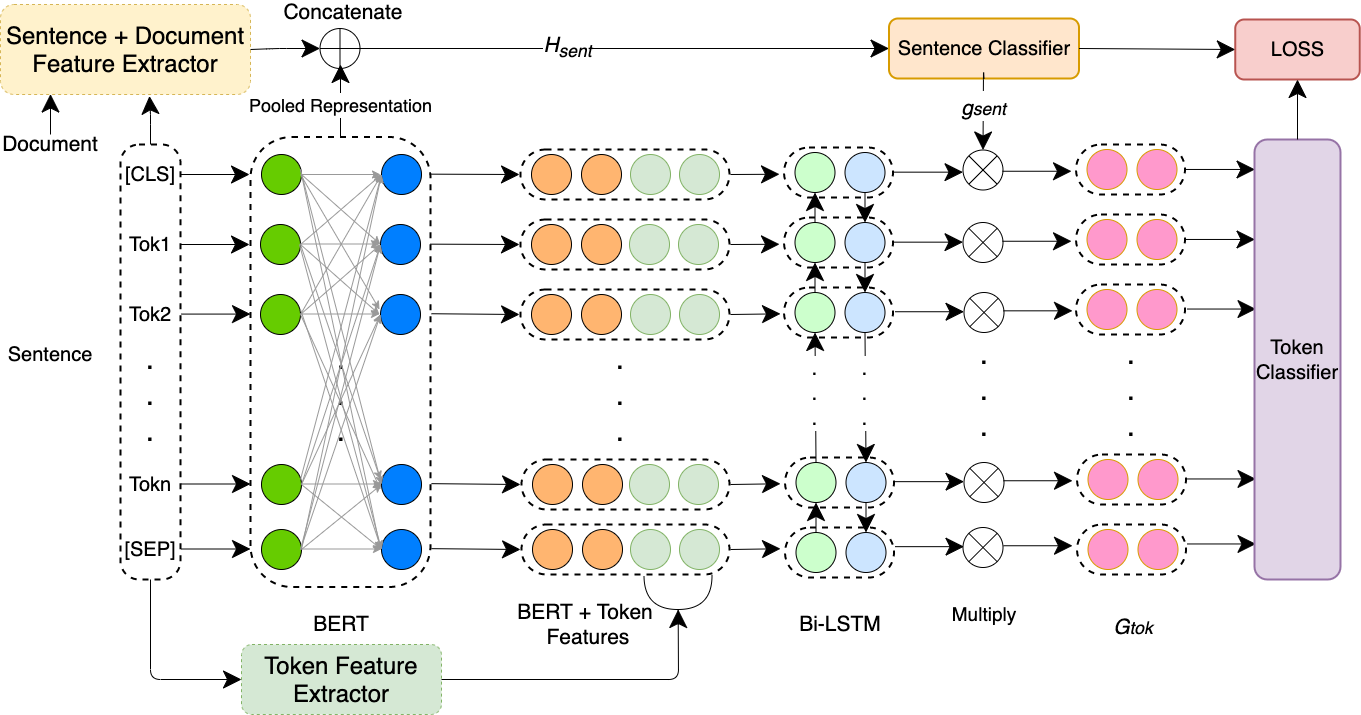}
\caption{Overview of our Multi-granular BERT-BiLSTM model. The sentence is passed into BERT and the token representations are concatenated with token features before being passed to a BiLSTM layer. The pooled representations are concatenated with sentence and document level features before getting the logits of the sentence containing propaganda. This gating value is then multiplied with the BiLSTM outputs of each token before being used to predict token labels. The model is jointly trained using loss from both token and sentence classification (Details in Section
~\ref{sec:details}). 
}
\label{fig:multi-granular}
\end{figure*}

\section{Model}
\label{sec:details}
In this section, we describe the modeling aspects we have adopted for the task. The overview of our model is given in \autoref{fig:multi-granular}.


\subsection{Multi-granular BERT-BiLSTM}
We implement a model that leverages sentence-level propaganda detection to guide the task of detecting propaganda word-spans \cite{propaganda-emnlp-2019}. Both stages have their separate classification layers $L_{sent}, L_{tok}$ and the feature representation learnt at the sentence-level $H_{sent}$ influences $H_{tok}$.

To classify each sentence, we mean-pool the token representations output from BERT, concatenate it with the sentence ($\bm{F}_{sent}$) and document ($\bm{F}_{doc}$) level features, and pass it through a fully-connected layer to obtain the logits $\bm{p}_{sent}$. $[; ]$ denotes the concatenation operator. 
\begin{equation}
    \begin{split}
        \bm{H}_{sent} &= [\operatorname{Pool}(\operatorname{BERT}(\small{[\text{CLS}]}, w_1, w_2, \ldots, w_n, \small{[\text{SEP}]})); \bm{F}_{sent}; \bm{F}_{doc}], \hspace*{1cm} \bm{p}_{sent} = \operatorname{FC}(\bm{H}_{sent}).
    \end{split}
\end{equation}

For word-span level predictions, we input the token representations output from BERT and concatenate them with word-level features for each token ($\bm{F}_{word}$) before passing them into a BiLSTM: 

\begin{equation}
    \begin{split}
        \bm{H}^*_{tok} &= [\operatorname{BERT}(\small{[\text{CLS}]}, w_1, w_2, \ldots, w_n, \small{[\text{SEP}]}) \oplus \bm{F}_{word}], \hspace*{1cm} \bm{H}_{tok} = \operatorname{BiLSTM}(\bm{H}_{tok}^*).
    \end{split}
\end{equation}


where $\oplus$ represents the token-wise concatenation operator
and  $\bm{H}_{tok}$ is the list of contextual representations for each token output by the BiLSTM. $\bm{F}_{word}, \bm{F}_{sent},$ and $\bm{F}_{doc}$ are described in Section~\ref{sec:add}. 

The sentence-level representation is passed through a masking gate that projects $\bm{p}_{sent}$ to a single dimension and applies sigmoid non-linearity. To control the information flow, each element of the token-level representation $\bm{H}_{tok}^i$ is multiplied with the gate to obtain gated token representations $\bm{G}_{tok}$.
This biases the token-level model to ignore samples strongly classified as negative in the sentence-level task, allowing it to selectively learn additional information to improve performance on token-level prediction:
\begin{equation}
    \begin{split}
        \bm{g}_{sent} = \sigma(W_g~\bm{p}_{sent} + b_g), \hspace*{1cm}
        \bm{G}_{tok}^i = \bm{g}_{sent} * \bm{H}_{tok}^i, \hspace*{1cm} \bm{p}_{tok}^i = \operatorname{FC}(\bm{G}_{tok}^i) .\\
    \end{split}
\end{equation}

We use a cross-entropy loss with sigmoid activation for sentence-level classification, whereas the token-level classification is trained using a cross-entropy loss with softmax activation. The losses $L_{sent}, L_{tok}$ are jointly optimized using hyper-parameter $\alpha$ to control the strength of both losses: 
\begin{equation}
    \bm{L} = \alpha \bm{L}_{sent} + (1 - \alpha)\bm{L}_{tok} .
\end{equation}
\vspace{-10mm}
\subsection{Additional Features}
\label{sec:add}
We incorporate additional lexical, syntactic, linguistic, and topical features at word, sentence and document levels to better inform the model.

\label{sec:prop_feats}

\noindent{\textbf{Syntactic Features:}}
Inspired by prior work that leveraged syntactic features effectively  \cite{reside,amused}, and motivated by the observation that many propaganda spans are persuasive or idiomatic phrases, we extract phrasal features from constituency parse trees to explicitly  incorporate structural syntactic information. We encode the path from a word to the root in the parse tree as a $d_c$-dimensional embedding.\footnote{We set the value of $d_c$ to $30$ in our experiments.} Stanford CoreNLP \cite{stanford_corenlp} was used to extract the constituency parses as well as part-of-speech tags that were also used as features. 

\noindent\textbf{Affective and Semantic Features}:
In addition to structural cues, prior work has shown that propaganda is marked with affective and emotional words and phrases \cite{linguistic-features-prop,uby-embeddings-propaganda}. Motivated by that, we append to word embeddings of the $i^{th}$ token (after BERT) features extracted using affective lexicons, including the LIWC \cite{liwc} dictionary, NRC lexicons of Word Emotion, VAD and Affect Intensity \cite{NRC-Emotion,NRC-VAD,NRC-AffectIntensity}. We also assign a score to each token that corresponds to the frequency of the word in the propaganda spans as opposed to the non-propaganda ones. For example, words like `invader', will have a high score, as it is salient to propaganda spans. Furthermore, we incorporate semantic class features. These include named entities (such as Person, Place, Temporal Expressions) as identified by the CoreNLP NER tagger \cite{stanford_corenlp} and finer-grained topical categories from Empath \cite{fast2016empath} (such as Government, War, Violence).

These word-level features ($\bm{F}_{word}$) are concatenated to the token-level BERT representations.

\noindent{\textbf{Sentence-level Features}} ($\bm{F}_{sent}$):
We also encode a sentence using the BERT-large-case model fine-tuned on news articles from different sources to get a 1024 dimensional vector.

\noindent{\textbf{Document-level Features}} ($\bm{F}_{doc}$):  Similarly, we obtain a 1024-dimensional embedding of the document, by averaging across BERT embeddings obtained for each sentence in the document.

The motivation for incorporating sentence and document features is to inform the model about the overall topical content of the article. We hypothesize that these features are especially helpful in detecting a specific type of propaganda called ``repetition'' where certain events are mentioned several times in the document. We append these to the pooled BERT representation to improve sentence-level classification.

\subsection{Unsupervised Fine-tuning}
\label{sec:prop_ft}
Neural language models like BERT~\cite{bert-devlin-2019} unlock their real power from the large amounts of data they are pre-trained on. The original BERT models are trained on BookCorpus and Wikipedia datasets that are essentially a text-based knowledge source. Therefore, the vanilla BERT learns language properties from an objective dataset and thus missing the nuances of persuasive and metaphorical language extensively used in propaganda \cite{propaganda-emnlp-2019}. To alleviate this issue we pre-train the original BERT models on a large collection of news articles collected from different sources. To ensure equal representation, we scraped approximately 10k articles from propaganda websites mentioned in  \newcite{propaganda-emnlp-2019}, e.g., Lew Rockwell,
SHTFplan.com, 
and 10k articles from trusted non-propagandist sources like CNN 
and New York Times.\footnote{\url{https://www.lewrockwell.com/, https://vdare.com/, http://www.shtfplan.com/, https://www.cnn.com/, https://www.nytimes.com/}} 
We scraped articles from 2018 to mid-2019 to ensure that the articles follow the same topical distribution as that of the training articles. We provide the unlabeled data, split to sentences, and train BERT on masked LM and next sentence prediction losses~\cite{bert-devlin-2019}. We leverage the pipeline provided by the Huggingface transformers library.\footnote{\url{https://github.com/huggingface/transformers}}

\subsection{Class-Imbalance}
\label{sec:prop_ci}
Since our corpus suffers from class imbalance in both sentence classification and word-level span identification task, we associate higher weight to the loss incurred by samples of the minority class. Following~\newcite{khosla2018emotionx}, we calculate these weights as inverse of the normalized frequency of samples of each class and plug them into the cross-entropy loss. Namely $L = -\frac{1}{N} \sum_{n=1}^{N} w_n * \text{loss}_n$, where $w_n$ is the weight associated with the loss for each sample. 

\section{Experimental Setup}
\label{sec:setup}
We used the BERT model from hugging-face library which was then fine-tuned on the corpus. 
Since the test-set was not made available during the competition, we hold out a small part of the training data as dev-set which was used to tune the models. The submitted model was chosen based on its Span-level Normalized F1-Score on the validation set provided by the competition organizers.\footnote{The official competition metric. We use the script provided by the organizers to calculate Span-level F1.} The hyperparameter choices are provided in Table~\ref{tab:hp} in the Appendix. We run all our experiments on Nvidia Geforce GTX 1080 Ti GPUs with 12GB memory.

\section{Results}
\label{sec:prop_results}
In this section, we present our method's results on the official validation set. Table~\ref{tab:embed} shows the performance of different multi-granularity BERT and BERT-BiLSTM. We report the mean of 5 independent runs (with different seeds). We find that BERT-large variants perform significantly better than BERT-base with BERT-large-cased performing the best. This suggests that case-based signals might be important for identifying spans as the writer might use capitalization to put more emphasis on propaganda information.


We also analyze the importance of word, sentence and document-level features (as detailed in Section \ref{sec:prop_feats}) to our model. For brevity, we only present the results on multi-granularity BERT-large-uncased BiSLTM (\textbf{MGU-LSTM}) and BERT-large-cased BiLSTM (\textbf{MGC-LSTM}) models due to their superior performance. Initial few rows in Table~\ref{tab:feats} depict the results for concatenating word-level features to word-level BERT representations. We find that concatenating Affective, LIWC, (\textbf{A}) and Syntax (\textbf{X}) features increases the performance of \textbf{MGU-LSTM} and \textbf{MGC-LSTM} models by $0.70$ and $0.24$ F1 respectively. Further adding NER and Empath (\textbf{N}) based features only seems to help \textbf{MGC-LSTM} attaining $0.18$ additional points.
This suggests that the embeddings learnt by the two models might differ in the kinds of features they represent. \textbf{MGU-LSTM} might already encode most of the information present in C (dimensions highly correlated with C). Nevertheless, testing this hypothesis requires an in-depth analysis of individual features and the word-level representations which is out of scope for this paper.

Our experiments with sentence (\textbf{S}) and document-level (\textbf{D}) features (Table~\ref{tab:feats}) suggest no clear pattern as adding sentence-level features to \textbf{MGC-LSTM-AXN} increases F1 by $0.1$ but degrades \textbf{MGU-LSTM-AX}'s performance from $44.11$ to $44.07$ F1. However, adding document-level features seem to benefit \textbf{MGU-LSTM-AX-S} but show no significant improvement ($p = 0.05$) in \textbf{MGC-LSTM-AXN}.

Finally, we present the contributions of the weighted cross-entropy loss (\textbf{W}) and the unsupervised fine-tuning (\textbf{F}) performed on articles from propaganda news sources in Table~\ref{tab:ft}. Adding inverse-frequency based weighting gives a boost of $0.66$ F1 points to \textbf{MGC-LSTM-AXN-S}. This is expected as the corpus is highly imbalanced towards non-propaganda words. Fine-tuning on news articles also provides a significant performance jump indicating the advantages of learning domain-informed contextualized representations. Our best performing (single) model is a multi-granularity BiLSTM architecture with BERT-large-cased embeddings fine-tuned on news articles, incorporating Affect, LIWC, Syntax, NER, Empath and sentence features, and optimized using a weighted cross-entropy loss.

\begin{minipage}[h]{0.5\textwidth}
    \vspace{0.2cm}
    \resizebox{\textwidth}{!}{
    \centering
    \begin{tabular}{lr}
        \toprule
        \textbf{Model} & \textbf{Span-level F1} \\
        \midrule
        BERT-base-uncased (BERT-BU) & $41.38$ \\
        BERT-base-cased (BERT-BC) & $42.12$ \\
        BERT-large-uncased (BERT-LU) & $43.12$ \\
        BERT-large-cased (BERT-LC) & $\bm{43.56}$ \\
        \midrule
        BERT-LU BiLSTM (\textbf{MGU-LSTM}) & $43.41$ \\
        BERT-LC BiLSTM (\textbf{MGC-LSTM}) & $\bm{43.86}$ \\
        \bottomrule
    \end{tabular}
    }
    \captionsetup{type=figure,font=small,labelfont=small}
    \captionof{table}{Multi-granularity model variants. The BERT based models are the baselines.}
    \label{tab:embed}
    \vspace{1.5cm}
    \centering
    \begin{tabular}{lr}
        \toprule
        \textbf{Model} & \textbf{F1} \\ 
        \midrule
        \textbf{MGC-LSTM-AXN-S} & $44.38$ \\
        + Weight (\textbf{W}) & $45.04$ \\
        + Fine-tune (\textbf{F}) & $44.56$ \\
        + \textbf{W} + \textbf{F} & $\bm{45.45}$ \\
        \bottomrule
    \end{tabular}
    \captionsetup{type=figure,font=small,labelfont=small}
    \captionof{table}{Results for Weighted cross-entropy loss (W) and unsupervised fine-tuning (F) on news articles.}
    \label{tab:ft}
    \vspace{1cm}
\end{minipage}
\hspace{1cm}
\begin{minipage}[h]{0.4\textwidth}
    \vspace{0.2cm}
    \centering
    \resizebox{\textwidth}{!}{
    \smaller
    \begin{tabular}{lr}
        \toprule
        \textbf{Features} & \textbf{F1} \\
        \midrule
        \multicolumn{2}{c}{\textbf{Word}} \\
        \midrule
        \textbf{MGU-LSTM} & $43.41$ \\
        + (Aff + LIWC) (\textbf{A}) &$43.69$ \\
        + \textbf{A} + Syn (\textbf{X}) & $\bm{44.11}$ \\
        + \textbf{AX} + (NER + Emp) (\textbf{N}) & $43.81$ \\
        \midrule
        \textbf{MGC-LSTM} & $43.86$ \\
        + (Aff + LIWC) (\textbf{A}) & $43.87$ \\
        + \textbf{A} + Syn (\textbf{X}) & $44.10$ \\
        + \textbf{AX} + (NER + Emp) (\textbf{N}) & $\bm{44.28}$ \\
        \midrule
        \multicolumn{2}{c}{\textbf{Sentence}} \\
        \midrule
        \textbf{MGU-LSTM-AX} + Sent (\textbf{S}) &$44.07$ \\
        \textbf{MGC-LSTM-AXN} + Sent &$\bm{44.38}$ \\
        \midrule
        \multicolumn{2}{c}{\textbf{Document}} \\
        \midrule
        \textbf{MGU-LSTM-AX-S} + Doc (\textbf{D}) &$44.27$ \\
        \textbf{MGC-LSTM-AXN-S} + Doc &$\bm{44.35}$ \\
        \bottomrule
    \end{tabular}
    }
    \captionsetup{type=figure,font=small,labelfont=small}
    \captionof{table}{Results for different word, sentence and document-level features. We only show the scores on MGU-LSTM and MGU-LSTM here for brevity (\textbf{Aff} = Affective features, \textbf{Syn} = Syntactic features, \textbf{Emp} = Empath features).}
    \label{tab:feats}
    \vspace{0.5cm}
\end{minipage}

\noindent\textbf{Ensemble}
We created a majority-voting ensemble of a subset of the model variants discussed above. We observe that we obtain the best results when the ensembled models are the most dissimilar. This pattern has also been shown to be beneficial in stacking where the same machine-learning problem is tackled with different types of learners. We find that an ensemble of 7 models namely
\textbf{BERT-LC-AXN-S-D-W} (seed 3), 
\textbf{MGC-LSTM-AX-S-W} (seed 3),
\textbf{MGU-LSTM-AX-S-W} (seed 1),
\textbf{MGC-LSTM-AXN-S-W} (seed 3),
\textbf{MGC-LSTM-AXN-S-W-F} (seed 1),
and \textbf{MGC-LSTM-AXN-S-W-F} (seed 3)
performs substantially better than individual models improving the performance on validation-set to $\bm{48.86}$. We illustrate an example of detecting propaganda spans by our model in Appendix. The spans are highlighted in bold. We observe that our model is not only able to detect short spans like  loaded language or named calling but also large spans corresponding to repetition and slogans that rely on contextual information.

\noindent\textbf{Final Submission}
We submit the results from the ensemble followed by post-processing like merging disjoint spans separated by by 1-2 words, removing stop-words or stray characters like quotes from the beginning and end of spans, as well as label words frequently labelled with the 'loaded language' label. Our final submission scored $\bm{49.06}$ F1-Score on the official dev-set and $\bm{47.66}$ on the official test-set.

\section{Conclusion}
\label{sec:conclusion}
This paper describes our $4^{th}$ place submission to the Span Identification (SI) subtask of SemEval 2020 Task 11. Our approach is based on a multi-granularity BiLSTM with BERT embeddings. We explore the contributions of several word, sentence and document-level features concatenated with token and pooled embeddings output from BERT. Our work also highlights the importance of tackling class-imbalance in the corpus and learning domain-informed representations through unsupervised fine-tuning of BERT on latest news articles. We submit a majority-voting ensemble of multiple models with potentially dissimilar decision boundaries to make robust predictions on the official validation and test-set.

\bibliographystyle{coling}
\bibliography{references}

\setcounter{table}{0}
\renewcommand{\thetable}{A\arabic{table}}
\appendix
\section{Hyper-parameters}
\begin{table}[h!]
    \centering
    \small
    \begin{tabular}{cc}
        \toprule
        \textbf{Hyper-parameter} & \textbf{Values} \\ 
        \midrule
        batch-size & $8,16$ \\
        epochs & $20$ \\
        learning-rate & $3e-5$ \\
        $\alpha$ & $0.9, 0.5$ \\
        l2 regularization ($\beta$) & $0.0001$ \\
        early-stopping patience & $9$ \\
        early-stopping criterion & Span-level F1 \\
        seeds & $1,2,3,12,123,1234,12345$ \\
        \bottomrule
    \end{tabular}
    \captionof{table}{Hyper-parameters}
    \label{tab:hp}
\end{table}

\section{Example of propaganda detection}

Tudor Simionov was killed when he tried to prevent a gang from \textbf{gatecrashing} a party in Mayfair. 
The Page One headline in the Evening Standard said it all: `First Stab Victims Of The New Year.' Welcome to 2019. ...
He died at the scene, in Park Lane, after being attacked by up to ten men — including, allegedly, the 26-year-old son of \textbf{Finsbury Park firebrand Captain Hook}, the \textbf{Islamist hate preacher} currently serving life ...
But a visible police presence may have deterred the gang who killed Tudor Simionov. ...
\textbf{No doubt London’s two-bob chancer of a Mayor Sadiq Khan would blame the absence of police }on the \textbf{streets on ‘austerity’ or the ‘Tory cuts’.}
Maybe if he hadn’t  \textbf{splurged £2.3 million of taxpayers’ money on a vainglorious New Year fireworks display,} he could have diverted more money to the Met to put extra bobbies on the beat.
Some of this largesse was frittered away turning the London Eye into a blue and yellow EU flag, as part of a \textbf{cynical, self-serving propaganda stunt aimed at bolstering the Stop Brexit campaign and burnishing the image of the Mayor himself.}...
What’s most likely to deter immigrants from coming here is not Khan slandering Leave voters as \textbf{knuckle-scraping, racist bigots,} but the news that someone can \textbf{be stabbed to death on one of London’s busiest, best-known and most upmarket streets} .... 
And if that \textbf{chilling reality} doesn’t put people off moving here from overseas, then perhaps they may be dissuaded by the \textbf{paralysis on London’s roads} ... 
\textbf{Rough sleepers and aggressive beggars} are a permanent fixture in the West End.
\textbf{Yet while innocent blood runs in the gutters,} and police budgets are restrained, Khan indulges in \textbf{fatuous propaganda stunts,} ... 
\textbf{London’s evening paper has a splash headline marking the first stabbings of the year.}

The first of many more to come, no doubt.
\textbf{Makes you proud to be British.
}


\end{document}